\documentclass[conference]{IEEEtran}
\IEEEoverridecommandlockouts
\usepackage{cite}
\usepackage{amsmath,amssymb,amsfonts}
\usepackage{algorithmic}
\usepackage{graphicx}
\usepackage{comment}
\usepackage{textcomp}
\usepackage{xcolor}
\usepackage{color}
\usepackage{lipsum}
\usepackage{times}
\usepackage{textcomp}
\usepackage{amsmath,amssymb,amsfonts}
\usepackage[inline]{enumitem}
\usepackage[ruled,vlined,linesnumbered]{algorithm2e}
\usepackage{color}
\usepackage{tablefootnote}
\usepackage{mathtools}

\usepackage{amsthm}
\usepackage{imakeidx}
\usepackage{float}
\usepackage{multicol}
\usepackage{booktabs} 
\usepackage{subcaption}
\usepackage{listings}
\usepackage{verbatim}
\usepackage{soul,xcolor}
\usepackage{xltabular}
\usepackage{makecell}
\usepackage{adjustbox}
\usepackage{array}

\makeatletter

\def\BibTeX{{\rm B\kern-.05em{\sc i\kern-.025em b}\kern-.08em
    T\kern-.1667em\lower.7ex\hbox{E}\kern-.125emX}}
\begin{document}

\title{OptABC: an Optimal Hyperparameter Tuning Approach for Machine Learning Algorithms}

\author{\IEEEauthorblockN{Leila Zahedi$^{1,2}$, Farid Ghareh Mohammadi$^{3}$, and M. Hadi  Amini$^{1,2,*}$}
\IEEEauthorblockA{\textbf{1} \textit{Knight Foundation School of Computing and Information Sciences,  Florida International University, Miami, FL, USA} \\
\textbf{2} \textit{Sustainability, Optimization, and Learning for InterDependent networks (solid) laboratory, FIU Miami, FL, USA}\\ \textbf{3} \textit{Department of Computer Science, University of Georgia, Athens, GA, USA}\\
 }
}
\maketitle
\begin{abstract}
Hyperparameter tuning in machine learning algorithms is a computationally challenging task due to the large-scale nature of the problem. In order to develop an efficient strategy for hyper-parameter tuning, one promising solution is to use swarm intelligence algorithms. Artificial Bee Colony (ABC) optimization lends itself as a promising and efficient optimization algorithm for this purpose. However, in some cases, ABC can suffer from a slow convergence rate or execution time due to the poor initial population of solutions and expensive objective functions. To address these concerns, a novel algorithm, OptABC, is proposed to help ABC algorithm in faster convergence toward a near-optimum solution. OptABC integrates artificial bee colony algorithm, K-Means clustering, greedy algorithm, and opposition-based learning strategy for tuning the hyper-parameters of different machine learning models. OptABC employs these techniques in an attempt to diversify the initial population, and hence enhance the convergence ability without significantly decreasing the accuracy. In order to validate the performance of the proposed method, we compare the results with previous state-of-the-art approaches. Experimental results demonstrate the effectiveness of the OptABC compared to existing approaches in the literature.
\end{abstract}

\begin{IEEEkeywords}
Automated Machine Learning, Automated Hyperparameter Tuning, Artificial Bee Colony Algorithm, Evolutionary Optimization
\end{IEEEkeywords}

\section{Introduction}
\subsection {Overview}
Nature Inspired Optimization (NIO) and Machine Learning (ML) are two prominent and momentous sub-fields of Artificial Intelligence (AI).
ML algorithms use data through experience and improve automatically, to solve different problems and generate knowledge. Recently, NIO has garnered much attention to solve such problems efficiently and effectively. These algorithms includes but not limited to Genetic Algorithms (GAs) \cite{schmitt2001theory}, Particle Swarm Optimization (PSO) \cite{kennedy1995particle}, Ant Colony Optimization (ACO) \cite{dorigo2006ant}, and Artificial Bee Colony (ABC) \cite{karaboga2005idea}. 
Among these approaches, ABC has been widely used in the literature due to its strong global search capabilities and a low number of parameters as compared with other nature-inspired algorithms. Therefore, it has been utilized in a wide range of applications to solve complex optimization problems \cite{dokeroglu2019artificial, mohammadi2021evolutionary, mohammadi2020applications, gao2018improved}.

ABC is a swarm intelligence method in which different  solutions are called food sources. The initial set of solutions is generated merely based on random distribution.
In ABC, three different types of bees use their search strategy to achieve new food sources. \textbf{Employed} and \textbf{Onlooker} bees have similar search strategies and are responsible for exploitation, while \textbf{Scout} bees are responsible for exploring and inserting new solutions into the population. 

Some of the previous studies report on the slow convergence of the primary ABC method or getting stuck to local optima \cite{shi2017balanced}. According to the literature, many of the ABC algorithms ignore the role of population initialization \cite{babaeizadeh2017enhanced}. Therefore methods that can provide richer populations can be very beneficial regarding both convergence rate and finding better solutions.

In the context of hyper-parameter tuning of ML algorithms, assuming that there are $M$ combinations and $P$ hyper-parameters per combination, we can construct an $M\times P$ matrix. In case $M$ and $P$ are too large, it is too expensive run the ML model on all the possible configurations. Additionally, ML algorithms have different objective functions depending on the complexity of the models themselves. For instance, long training time especially for large datasets is actually one of the SVM disadvantages \cite{ying2016improved}, because of its costly objective function.
Support Vector Machines (SVMs) are supervised ML models that utilize associated learning algorithms to detect patterns existing in data. SVM applies to regression, classification, and outlier detection problems. Taking the training data marked with their belonging classes, SVM creates a classifier using a hyper-linear separating plane and builds a model that predicts the classes of new data points. 

The high time complexity of SVM models (especially when working with large datasets) indicates expensive objective functions in the ABC algorithm as well, which leads to high execution time in comparison to other ML models \cite{zahedihyp}.
Random Forest and eXtreme Gradient Boosting (XGBoost) models are also among ensemble supervised machine learning models that use enhanced bagging and gradient boosting, respectively. These two models have proved a powerful predicting ability in different applications. However, they have more number of main hyper-parameters in comparison to other ML models which makes a large search space and hence increase the time complexity of ABC algorithm.

In this study, we propose a modified version of the Hyp-ABC algorithm in a previous work \cite{zahedihyp} called OptABC, to tune the main hyper-parameters of the mentioned ML algorithms; The K-means clustering algorithm is used to offer heterogeneity to the population of solutions. This method can enhance the convergence rate by avoiding the evaluation of all solutions in the population by taking only the cluster centroids and evaluating them. Moreover, we add an Opposition-Based Learning (OBL) method to random food source search in the original scout phase to discover richer and unvisited food source positions to improve the balance in exploration and exploitation steps.
The MIDFIELD dataset is used to verify the performance of OptABC in the experiments.

\subsection{Motivation}
ML model predictions are proven to be effective for accurate decision makings. However, to make a model work at its best, its hyper-parameters must be tuned. In Hyper-Parameter Optimization (HPO) problems, the goal is to build a model with the best set of hyper-parameters for an ML model to minimize the objective function or maximize the accuracy \cite{sun2019survey}:
\begin{equation}
x=arg \min_{x \in S} f(x)
\end{equation}
where $f(x)$ is the objective function or error rate that should be minimized, $x$ is the optimal set of hyper-parameters, $S$ is the search space, and $arg \min{f(x)}$ is the optimal set for which $f(x)$ reaches its minimum.

However, searching through a large number of configurations of hyper-parameters can be computationally highly expensive \cite{zahedi2021search}. Many of the HPO problems are non-convex optimization problems, meaning that they have multiple local optimums. Therefore, traditional optimization approaches are not a good fit for them \cite{luo2016review}. Recently, evolutionary optimization techniques have gained much success in different applications to solve non-convex complex optimization problems \cite{yao2018taking}. These techniques do not guarantee to find the global optimum; however, they detect near to global optimum within a few iterations.

Existing automated hyper-parameter tuning techniques suffer from high time complexity for some of the ML models, especially SVM \cite{zahedihyp} and tree-based models. This is due to the high time complexity of the SVM objective function, especially for large size datasets \cite{mulay2010intrusion}, and or the number of hyper-parameters in the search space.

This paper proposes OptABC, an automated novel hybrid hyper-parameter optimization algorithm using the modified ABC approach, to measure the ML models' classification accuracy. The main focus of this study is on the ABC algorithm. ABC operates based on the foraging behavior of honeybees and was first developed by Karaboga \cite{karaboga2005idea}. Previous studies show that the performance of the ABC algorithm is competitively better than that of other population-based algorithms \cite{sahoo2017two}. Thus, it has been used to solve different problems in different applications. As mentioned in the previous section, slow convergence because of the stochastic nature \cite{kang2013artificial} and sticking in the local optima for complex problems \cite{chang2009genetic} are of the disadvantages of the ABC algorithm. The challenge is exacerbated when dealing with expensive objective functions or large search spaces for evaluating the quality of the food sources.

In this paper, an effort has been made to address
the challenges mentioned above to make the HyP-ABC algorithm \cite{zahedihyp} more efficient for ML algorithms to solve optimization problems.

\subsection{Research Focus and Contribution}
In this research study, we mainly focus on the slow convergence issue of the ABC algorithm. This issue can be derived from the shortcomings of the ABC algorithm and the complexity of ML models.
These shortcomings are summarized as follows: Initial position of food sources, the position of the food sources by scout bees, and expensive evaluation function in the foraging process. To address these concerns, we propose an improved version of a previous ABC-based algorithm. The contribution of this paper is threefold as follows:
\begin{enumerate}
    \item Presents an automated hyper-parameter tuning method for different ML models using evolutionary optimization for large real-world datasets.
    \item Presents a K-Means clustering approach to form a better initial population.
    \item Determines the abandoned food source position by applying the OBL method in addition to the random search strategy to strengthen the exploration phase of ABC.
\end{enumerate}

The proposed algorithm is evaluated on a real-world educational dataset, and the results are compared with an algorithm in a previous study in 2021\cite{zahedihyp}. The proposed method provides better performance in terms of execution time without decreasing the accuracy in most of the cases.

\subsection{Organization} The rest of this paper is organized as follows. Section \ref{sec:RelatedWork} highlights the background and related work regarding ABC algorithm and variants of ABC in different applications. Section \ref{sec:Approach} elaborately explains our novel approach. Section \ref{sec:methodology} and \ref{sec:Results} presents our experimental methodology and results, followed by section \ref{sec:Conclusion}, that concludes the paper.

\section{BACKGROUND REVIEW AND RELATED WORK}
\label{sec:RelatedWork}
This section discusses the background review of
this work, including the basic ABC algorithm, its different steps, and ABC-related works in different applications.
\subsection{Overview of ABC Algorithm}
ABC algorithm works based on the natural foraging process of honey bees swarm and was first introduced by Karaboga in 2005 \cite{karaboga2005idea} to solve optimization problems. As mentioned in previous section, this algorithm consists of three different phases. Initialization and exploration is performed by scout bee; the Scout bee randomly pick the food sources, then each food source is assigned to the Employed bee for exploitation. Afterward, the Onlooker bees wait in the hive for the information that the Employed bees share with them. Once they received the information, they may exploit or abandon the food sources based on their quality. The abandoned food source is replaced with a discovered random food source by the Scout bee. This process continues until the termination criterion is achieved. A summarized mechanism of the original ABC algorithm is given below:
\subsubsection{Initialization}
The initial population containing different food sources (vectors) using a random strategy is created using below formula:
Then each of the vectors is assigned to one employed bee.
\begin{equation} X_{i,j}=x_{min,j}+rand(0,1)(x_{max,j}-x_{min,j})
\label{eqn:init}
\end{equation}
Where $rand()$ returns a random number within the provided range.
\subsubsection{Employed Bee Phase}
Each of the Employed bees detect a neighbor food source, $V_i$, by equation \ref{eqn:employed}. This is done by changing only one of the dimension's values \(k\)th in the vector. Then, the employed bee compares $V_i$ with $X_i$ and selects the better vector.
\begin{equation} 
V_{i,j}=x_{i,j}+rand(-1,1)(x_{i,j}-x_{k,j})
\label{eqn:employed}
\end{equation}
Where $k$ and $i$ are different.

\subsubsection{Onlooker Bee Phase}
In the next phase, Employed bees share the information with onlooker bees. Then onlooker bees calculate the probabilities of vectors ($P_{i}$), based on roulette wheel selection in equation \ref{eqn:prob}. Onlooker bees select or leave the vectors based on the calculated probability values. The vector with a higher probability is more likely to get selected by the Onlooker bees.
\begin{equation} 
P_{i}=0.9\frac{Fit_{i}}{max(Fit)}+0.1
\label{eqn:prob}
\end{equation}
Where $Fit_{i}$ is the fitness value for the \(i\)th vector and is directly calculated from objective function, by equation \ref{eqn:fitness}.
\begin{equation}
fit_{i}=
\begin{cases}
\frac{1} {1+ f_{i}}, & \text{$f_{i}$ $\geq$ 0} \\
{1+ abs(f_{i})}, & \text{$f_{i}$ $<$ 0}
\end{cases}
\label{eqn:fitness}
\end{equation}
If an onlooker bee selects a vector, it generates a new vector solution similar to equation \ref{eqn:employed} and selects the better solution between $V_i$ and $X_i$. Next, the best vector of solutions visited thus far is memorized.

\subsubsection{Scout Bee Phase}
In the last phase, an exhausted food source (vector), if any, is determined by scout bee and exploration starts. In this phase, scout bee generate a random vector by equation \ref{eqn:scout} and replace it with the exhausted vector. This step is done without using any experience or greedy approach.
\begin{equation} X_{i,j}=x_{min,j}+rand(0,1)(x_{max,j}-x_{min,j})
\label{eqn:scout}
\end{equation}

The above phases repeat until the resource budgets, such as time or the maximum number of evaluations are exhausted, or when the desired results are achieved.

\subsection{Different Variants and Applications of ABC}

This section describes some of the related studies of ABC employed in various optimization problems. Researchers have used ABC to address feature selection problems and to speed up the classification process \cite{mohammadi2014image} and eliminate non-informative features \cite{mohammadi2014image, sarac2021artificial,mazini2019anomaly}. ABC has also been used to design automatically and evolve hyper-parameters of Convolutional Neural Networks (CNNs) \cite{zhu2019evolutionary}. 
Choong et al. improved the neighborhood search algorithm of the original ABC in the exploitation phases of the original ABC \cite{chang2018solving}. The experimental results of this study showed that the proposed function has some advantages over other search processes \cite{chang2018solving}. In \cite{mala2021hybrid,zhao2020improvement}, the authors proposed to modify the initialization and exploitation phases of ABC to improve the convergence. The authors tested the performance of their proposed algorithms and observed promising results. 

Pandiri and Singh \cite{pandiri2018hyper} developed a hyper-heuristic-based ABC algorithm for the k-Interconnected multi-depot multi-traveling salesman problem. The authors used an encoding scheme inside the algorithm. The study results showed that although the search space is smaller than previous schemes, it has better performance than previous methods applied in the literature.
Gunel and Gor used a dynamically constructed hyper-sphere to modify the ABC algorithm to improve the exploitation ability of ABC. In this method, the best food source was used to define the mutation operator. The solutions for the differential equations were obtained by training neural networks employing the modified ABC \cite{gunel2019modification}.

In another study, Agrawal introduced an improved version of the ABC using some features of the Gaussian ABC to improve the slow convergence of the original ABC. This approach was used to modify the employed, onlooker, and scout phases. The experimental results of this study showed the superiority of the method over the original ABC in the majority of the experiments \cite{agrawal2020modified}.

\subsection{Studies Employing Learning Techniques on ABC}
Many previous studies use hybridization approaches combining ML algorithms and nature-inspired algorithms to improve their performance regarding accuracy and time. These studies aim to provide more intelligent forms of ABC algorithm in different applications. This section covers some of the previous studies that used learning algorithms to enhance the optimization process of the ABC algorithm. The most common methods used in the literature are clustering, reinforcement learning, OBL.

Clustering has been used mostly in different ways in the initialization phase of the ABC algorithms. K-means clustering was either used to add diversity to the population or to detect the clusters in the optima \cite{sahoo2017two,ibrahim2020brain,zhou2019application}. Reinforcement learning has also been used in several studies to improve the searching process of the ABC \cite{zhao2020decomposition,fairee2019combinatorial}. This method is mostly used in the onlooker and employed phases of the ABC algorithm. OBL approach is another technique that has been used in previous studies to enhance the performance of the ABC method. This method has been employed in initialization \cite{sharma2020artificial,liu2019enhanced}, employed and onlooker phases \cite{sharma2018opposition,zhao2015artificial} .

\section{Proposed Approach}
\label{sec:Approach}
One of the disadvantages of ABC, like other population-based algorithms, is requiring a suitable initial population. Also, ABC sometimes suffers from a slow convergence rate \cite{zahedihyp}.
In some applications, such as ML problems, inserting a solution into the problem and testing its impact on the results to compute the fitness value may require a long time \cite{karaboga2020survey}. In such situations, learning-based models can be developed to guess the fitness values of the present solutions more quickly. 
Having a rich and diverse initial population can significantly improve the performance and convergence rate. Additionally, although ABC performs well at exploration, the Scout phase can still improve to guide the search process better and increase the likelihood of finding better food sources in the exploration process.

\begin{figure}
	\centering
		\includegraphics[width =0.445 \textwidth]{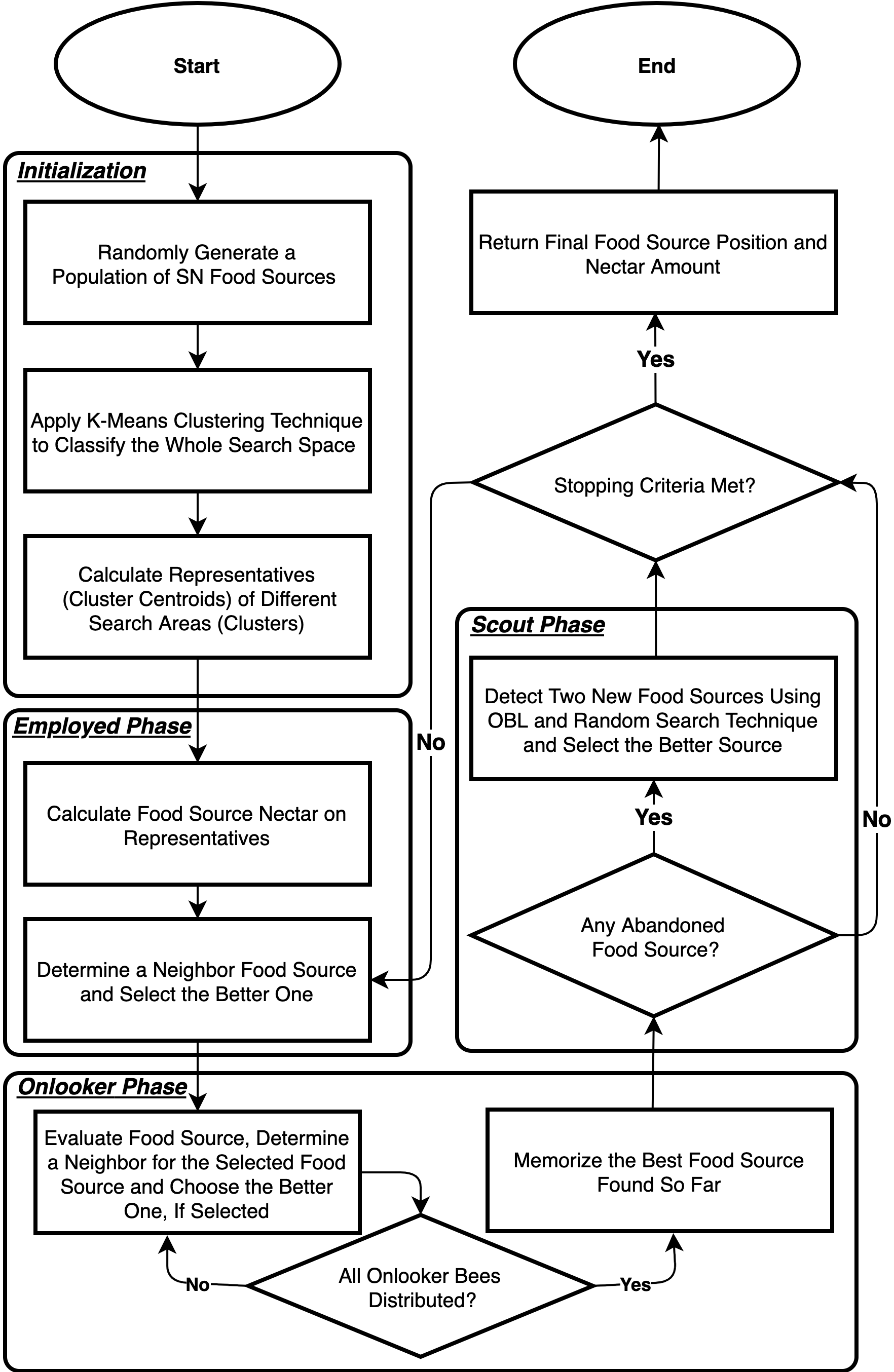}
	\centering
	\caption{Overall flowchart process}
	\label{fig:general}
\end{figure}

To address the above-mentioned inherent defects of ABC and accelerate the convergence rate of the algorithm, we propose an improved variant of an existing algorithm (HyP-ABC)\cite{zahedihyp} using learning algorithms. Employing the OptABC, main hyper-parameters of the three models are optimized. Therefore the optimal solution would be a vector with a dimension that equals to the number of main hyper-parameters.
OptABC applies three modifications. We employ K-Means clustering and Opposition-Based Learning (OBL) strategies in the initial population generation. The third modification is constructed in the scout bee search using the OBL method. 
Figure \ref{fig:general} shows the overall flowchart process of our paper.\\

The main structure of this novel algorithm, including two modified phases, can be summarized as below.

\begin{figure*}
	\centering
		\includegraphics[width =0.86 \textwidth]{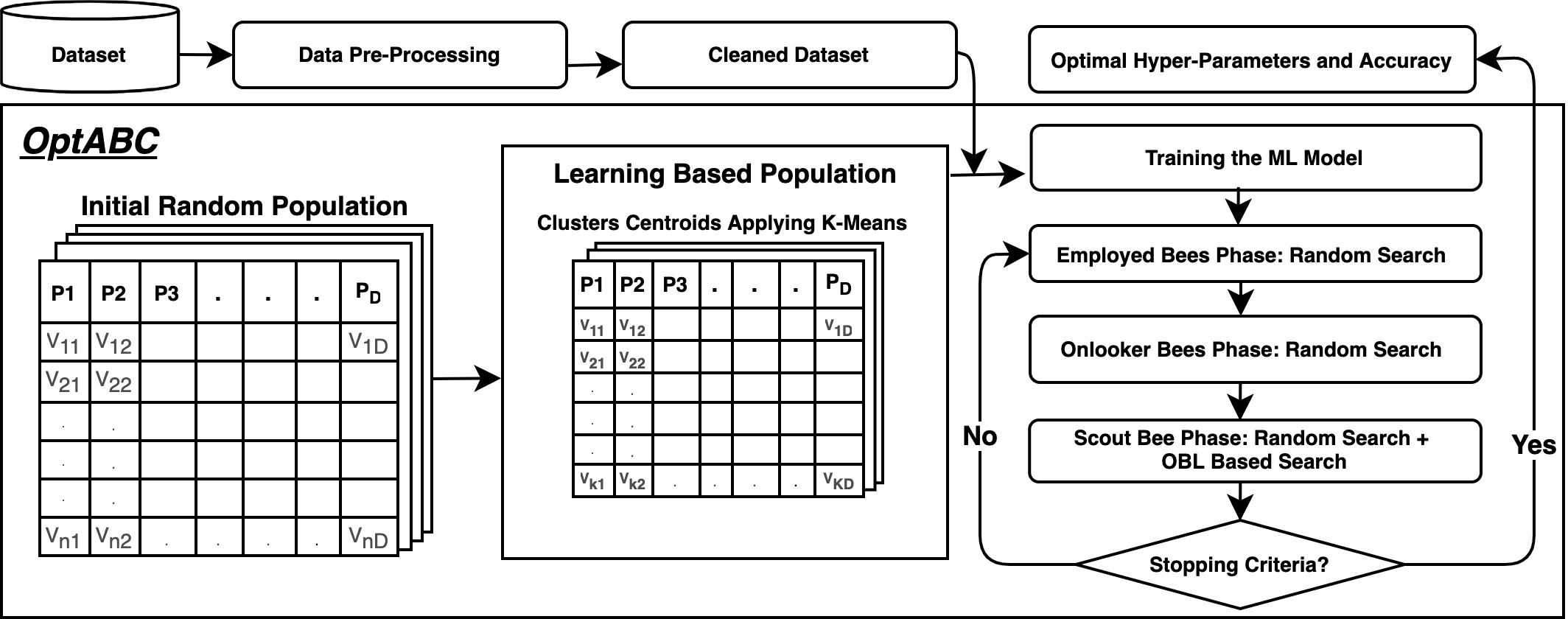}
	\centering
	\caption{Proposed Framework of OptABC}
	\label{fig:framework}
\end{figure*} 

\subsection{The Generation of Initial Population}
In HPO problems, each food source is a vector of hyper-parameters, $X_i,j$ by equation \ref{eqn:init}, where $j={1,2,...,D}$, $i={1,2,..,PN}$ and D and NP are the number of hyper-parameters and the number of population, respectively. Original ABC algorithm starts with generating randomly distributed food source locations. To improve the convergence rate of the ABC algorithm, we use a K-Means clustering algorithm \cite{macqueen1967some} to have a more diversified population. In this approach, we partition the randomly initiated population into $k$ clusters. Then a loop starts assigning the food sources to the closest mean, and the cluster centroids ($c^i$) are determined. This process continues until no change is observed in the position of cluster centers. Instead of calculating the objective function for all the food resources, the final clusters' centroids (by equation \ref{eqn:kmeansc}) are taken as representatives of each cluster for a new population. 
\begin{equation}
{\mu}_{i}=\frac{\sum_{j=1}^{PN}1\{c^i=j\}x^i}{\sum_{j=1}^{PN}1\{c^i=j\}} \hspace{30pt}j=1,2,...,k
\label{eqn:kmeansc}
\end{equation}
This approach helps to avoid evaluating expensive objective functions for each single food source in the initial population. The pseudo-code of the modified initialization step is described in Algorithm \ref{alg:kmeans}.

\begin{algorithm}
 \KwData{number of clusters ($k$) }
 Randomly generate a population of PN food sources\;
 Randomly select $k$ food sources from the generated population and assign them to each cluster\;
 \While{centroid position changes}{
  Assign each food source to its closest cluster\;
  Calculate new centroids (mean) of all clusters using equation \ref{eqn:kmeansc}\;
 }
 Take final $centroids$ as $population_{new}$\;
 \Return $population_{new}$
 \caption{Modified Initialization Phase (K-Means)}
 \label{alg:kmeans}
\end{algorithm}
\vspace{-5pt}

\subsection{Employed Bee Phase}
In this phase, employed bee discovers a new food source, $V_i$, in the vicinity of the current food source. In other words, only one hyper-parameter of the current food source changes in this phase. Then, the employed bee compares $V_i$ with $X_i$ and selects the better food source.
\begin{equation} 
V_{i,j}=x_{i,j}+rand(-1,1)(x_{i,j}-x_{k,j})
\label{eqn:emp-onl}
\end{equation}
\subsection{Onlooker Bee Phase}
Next, employed bees share the information with onlooker bees through waggle dance. Onlooker bees in the OptABC algorithm compute the probabilities of each food source. Then the probability of the food source is then compared to a random number between zero and one. Depending on the quality of the food source, onlookers may leave or exploit them. In this study, the probability of the food sources is calculated using min-max normalization in Equation \ref{eqn:prob2} ($0\leq P_{i} \leq1$, $min(Fit)\leq Fit_{i} \leq max(Fit)$). 
In other words, the range of all fitness values is normalized so that each value contributes approximately proportionately when it is compared to the random number. Then a comparison is made between the probability and random numbers for further decisions.

\begin{equation} 
P_{i}=\frac{Fit_{i}-min(Fit)}{max(Fit)-min(Fit)}
\label{eqn:prob2}
\end{equation}

If the food source gets selected for further exploitation, a new solution is generated using equation \ref{eqn:emp-onl}, and the algorithm goes on with the best solution.

\subsection{The Search Mechanism of Scout Phase}
In the Scout phase, if a food source does not get updated a defined number of times (limit), it will be considered an abandoned food source, and its assigned employed bee turns to a scout bee for discovering a new food source location. In original ABC, this is done by using equation \ref{eqn:scout}. As can be seen, the new food source is generated randomly and may not always be the best solution for an optimization problem. Hence, in this study, we consider operator adaptation to generate the new food source. In other words, in addition to detecting a new location for the abandoned food source using a random search technique, we also apply an OBL search technique.
OBL technique is a new concept in ML which was first introduced by Tizhoosh (2005) \cite{tizhoosh2005opposition} which is inspired from the opposition concept in the real world, Implying that Having a data point that is closer to an optimum point can result in faster convergence. However, if the optimum point is in the opposite direction (position), the search process needs more resources to find them. Therefore, searching in the opposites locations may help the algorithm converge faster. 
The opposition of each food source (a vector with D parameters) is calculated by equation \ref{eqn:scout2} to generate an alternative food source in the opposite location. This technique strengths the exploration phase and better guide the search process. In this phase, the food source with better quality gets replaced with the abandoned food source.

\begin{equation}
{\tilde x}_{i,j}={x}_{max,j} + {x}_{min,j} - {x}_{i,j}
\label{eqn:scout2}
\end{equation}

\subsection{Fitness Function of OptABC}
The fitness function for all the phases above in the proposed algorithm is calculated directly from the objective function (\ref{eqn:fitness}). The main goal in ML-related problems is to maximize the objective function, which is computed from performance indicators such as accuracy for classification problems or mean absolute error for regression problems. In this study, our goal is to maximize training objective function, which is accuracy-oriented and is calculated by the below formula:
\begin{equation}
Accuracy=\frac{TP+TN}{TP+TN+FP+FN}
\label{eqn:kmeans}
\end{equation}
Where $TP$, $TN$, $FP$, and $FN$ are true positives, true negatives, false positives, and false negatives, respectively. Since we are dealing with a balanced dataset we use accuracy as the overall performance metric.

\subsection{Framework and Time Complexity of OptABC}
The proposed framework of OptABC is given in Figure \ref{fig:framework}, and Algorithm \ref{alg:optABC} shows the previous work process\cite{zahedihyp}, along with the changes we made to improve the ABC convergence rate (in red).

\setstcolor{red}
\begin{algorithm*}
 \KwData{$population\_size$, $Search\_Space$}
 {Call Algorithm \ref{alg:kmeans}}\\

 \While{Stopping criteria is not met}{
 \For {$i$ to $population_{new}\_size$}{
  Employed bee generates a new food \(N{i}\) source in neighborhood and modify the new food source to an accepted food source \(M{i}\)\;
  Train the ML model with the updated food source\;}
  \For {$j$ to $population_{new}\_size$}{
  Onlooker bee calculate the probability ($P{i}$) of each solution based on Equation \ref{eqn:prob2}\\
  \If { \((rand{(0,1)})\) $>$ \((P{i})\) }{
  Onlooker bee generates a new food source in neighborhood \(N{i}\) and then modify the new food source to an accepted food source \(M{i}\)
  Train the ML model with the updated food source\;}
  \Else {Onlooker disregard the food source and moves to the next food source\;}
  }
  Memorize and update the best food source\;
  \If {trial $>$limit}{
  Scout bee generates a food source based on Equation \ref{eqn:scout}\;
 {Scout bee generates a food source based on Equation \ref{eqn:scout2}}\;
 {Select the food source with better quality}\;
  }}
\Return {Optimal food source}
\caption{OptABC Algorithm - Modifications}
\label{alg:optABC}
\end{algorithm*}

In this method, the number of initial solutions is $PN$, and the number of the secondary population is $\frac{PN}{K}$. The number of employed bees and onlooker bees is the same ($k$). In the employed bees phase, $k$ new solutions are generated. Similarly, another $k$ solutions are generated by onlooker bees. In the scout bee search phase, two solutions (One OBL-based solution and one random-based solution) are generated for each exhausted solution, and the best one is selected to get replaced with the exhausted food source. The random solution helps the algorithm escape from the local optimum, and the opposition solution can increase the probability of reaching better candidate solutions (compared to random search) \cite{rahnamayan2008opposition}. 

The time complexity of the proposed approach has not increased in comparison with the traditional ABC algorithm since not all solutions in the population are evaluated. ABC algorithm's time complexity depends on population size ($PN$), complexity of objective function ($f$) and maximum number of iterations ($I_{max}$) and number of clusters ($k$). Therefore, the time complexity of the proposed method is $O(I_{max}.(\frac{PN}{K}.f + \frac{PN}{K}.f + 2f))= O(I_{max}.(\frac{PN.f}{K}))$,
while the time complexity of original ABC is $O(I_{max}.(PN.f + PN.f + f))= O(I_{max}.(PN.f))$.
\section{Experimental Methodology}
\label{sec:methodology}
This section covers the methodology of our experiment. We leveraged our proposed methods to tune the main hyper-parameters of three ML models. As can be seen in figure \ref{fig:framework} the overall process is done in several phases, including data pre-processing and feature engineering, generating a new population, training SVM model, and ABC-based optimization steps. The Opt-ABC algorithm proposed in this study is an improved version of recent work in 2021\cite{zahedihyp}.

\subsection{Data Pre-Processing}
Data pre-processing in the context of ML refers to the technique of transforming the original data into an appropriate and understandable state for ML algorithms. A summarized mechanism of the steps for data pre-processing is given below:
\subsubsection{Data Cleansing}
In this step, the features with more than 60\% percent missing values are eliminated from the dataset. The target population in this study are only the student majoring in computing fields. Therefore, the data is filtered to include only students from computing majors.
\subsubsection{Feature Scaling}
Feature scaling is a technique to standardize the range of data, especially when there is a broad variation between values among different features to avoid biases from big outliers. We used \textit{StandardScaler} from the Scikit-learn package was used for scaling the input values.
\subsubsection{Feature Engineering}
In this step, one-hot encoding was utilized for encoding the categorical variables to binary representations.

\subsection{Hyper-Parameter Tuning}
Tuning hyper-parameters is a primary task in automated ML that helps the model achieve its best performance. Previous research reported on costly objective functions of ML models (specifically SVMs) regarding tuning time when using automated HPO methods \cite{zahedihyp}. Therefore Tree-based SVM and (RF and XGBoost) algorithms are selected to be explored in this study. The proposed learning-based improvements in the OptABC, are an attempt to provide the algorithm with a richer population and strengthen the algorithm's exploration phase.

\section{Experimental Results and Discussions}
\label{sec:Results}
This section presents the performance metrics, and experimental results after applying the OptABC HPO algorithm.

\subsection{Metric for Performance Evaluation}
The experiment is performed in two different stages; training using 80:20 ratio train-test sets and cross-validation. Regarding the experiments using cross-validation, the dataset is partitioned into three folds to evaluate the ML algorithms. Different folds are assigned to the training and test sets. Specifically, for each run, $f=1,2,3$, fold $f$ is assigned to the test, and the rest of the folds are assigned to the training set. Cross-validation is employed to reduce the chances of overfitting. Although since it has an impact on tuning time, we leverage parallel cross-validation. In each run, accuracy and the average overall accuracy over three runs were reported. We also use execution time as another performance metric to measure the time regarding running time to return the optimal set of hyper-parameters.

In this experiment, the \textit{Scikit-learn} libraries, along with other Python libraries, were used to leverage the ML models. All experiments were conducted using Python 3.8 on High-Performance Computational (HPC) resources.

\subsection{Performance Comparison of Different Population Sizes}
In this section, we compare the efficiency of OptABC with a previous method under different population sizes. The HyP-ABC \cite{zahedihyp} uses the random search strategy, while OptABC employs two different learning algorithms in the initial and the Scout phase of the algorithm to enhance the convergence rate of the previous method. The results are shown in Tables \ref{tab:RF}, \ref{tab:XGB}, and \ref{tab:SVM}. The running times in hours stand for the time for all iterations until getting the desirable accuracy.

\begin{table}[ht]
\centering
\caption{Performance evaluation of applying OptABC and HyP-ABC to the RF classifier on the MIDFIELD dataset.}
\label{tab:RF}
\def\arraystretch{1.5}
\resizebox{\columnwidth}{!}{%
\begin{tabular}{
  |c|c|c|c!{\vrule width 1.5pt} c|c|c|
  }
\hline
\multicolumn{7}{|c|}{\textbf{80:20 ratio Test-Train set}}\\
\hline
\textbf{} & \multicolumn{3}{c!{\vrule width 1.4pt}}{OptABC} & \multicolumn{3}{c|}{HyP-ABC}\\
\hline
\textbf{Population size} & \textbf{20} & \textbf{50} & \textbf{100} & \textbf{20} & \textbf{50} & \textbf{100} \\
\hline
\textbf{Accuracy(\%)} & 88.71 & 88.75 & 88.76 & 88.71 & 88.74 & 88.78\\
\hline
\textbf{\makecell{Execution Time(hrs)}} & \textbf{1.6} & \textbf{3.17} & 7.8 & 2.37 & 9.07 & \textbf{3.18}\\
\hline
\multicolumn{7}{|c|}{\textbf{With 3-fold cross validation}}\\
\hline
\textbf{Accuracy(\%)} & 88.79 & 88.77 & 88.77 & 87.57 & 88.74 & 88.77\\
\hline
\textbf{\makecell{Execution Time(hrs)}} & \textbf{0.46} &  \textbf{1.09} & 2.16 & \multicolumn{3}{|c|}{\textbf{$1.55^*$}}\\
\hline
\end{tabular}
}

\end{table} 

\begin{table}[ht]
\centering
\caption{Performance evaluation of applying OptABC and HyP-ABC to the XGBoost classifier on the MIDFIELD.}
\label{tab:XGB}
\def\arraystretch{1.5}
\resizebox{\columnwidth}{!}{%
\begin{tabular}{
  |c|c|c|c!{\vrule width 1.5pt} c|c|c|
  }
\hline
\multicolumn{7}{|c|}{\textbf{80:20 ratio Test-Train set}}\\
\hline
\textbf{} & \multicolumn{3}{c!{\vrule width 1.4pt}}{OptABC} & \multicolumn{3}{c|}{HyP-ABC}\\
\hline
\textbf{Population size} & \textbf{20} & \textbf{50} & \textbf{100} & \textbf{20} & \textbf{50} & \textbf{100} \\
\hline
\textbf{Accuracy(\%)} & 88.65 & 88.73 & 88.86 & 88.60 & 88.70 & 88.75\\
\hline
\textbf{\makecell{Execution Time(hrs)}} & \textbf{1.98} & \textbf{3.85} & \textbf{3.42} & 4.07 & 5.85 & 4.23\\
\hline
\multicolumn{7}{|c|}{\textbf{With 3-fold cross validation}}\\
\hline
\textbf{Accuracy(\%)} & 88.29 & 88.64 & 88.72 & 87.97 & 88.64 & 88.84\\
\hline
\textbf{\makecell{Execution Time(hrs)}} & \textbf{4.83} &  \textbf{19.54} & \textbf{17.6} & \multicolumn{3}{|c|}{$29.22^*$}\\
\hline
\end{tabular}
}

\end{table} 

\begin{table}[ht]
\centering
\caption{Performance evaluation of applying OptABC and HyP-ABC to the SVM classifier on the MIDFIELD dataset.}
\label{tab:SVM}
\def\arraystretch{1.5}
\resizebox{\columnwidth}{!}{%
\begin{tabular}{
  |c|c|c|c!{\vrule width 1.5pt} c|c|c|
  }
\hline
\multicolumn{7}{|c|}{\textbf{80:20 ratio Test-Train set}}\\
\hline
\textbf{} & \multicolumn{3}{c!{\vrule width 1.4pt}}{OptABC} & \multicolumn{3}{c|}{HyP-ABC}\\
\hline
\textbf{Population size} & \textbf{20} & \textbf{50} & \textbf{100} & \textbf{20} & \textbf{50} & \textbf{100} \\
\hline
\textbf{Accuracy(\%)} & 87.87 & 87.86 & 87.86 & 87.80 & 87.86 & 87.85 \\
\hline
\textbf{\makecell{Execution Time(hrs)}} & \textbf{18.3} & \textbf{8.61} & \textbf{15.5} & 23.95 & 53.05 & 62.57 \\
\hline
\multicolumn{7}{|c|}{\textbf{With 3-fold cross validation}} \\
\hline
\textbf{Accuracy(\%)} & 87.99 & 88.00 & 88.00 & 87.93 & 88.00 & 88.00 \\
\hline
\textbf{\makecell{Execution Time(hrs)}} & \textbf{19.71} &  39.01 & 40.10 & \multicolumn{3}{|c|}{\textbf{$34.15^*$}} \\
\hline
\end{tabular}
}

\end{table} 
As can be seen from the tables, in the majority of the cases, the execution time after applying OptABC algorithm has improved without decreasing the classification accuracy. 
In summary, our proposed algorithm is quicker than HyP-ABC. This difference is more tangible when the population is small showing the capability of the algorithm in finding the optimal solution when the population is not too large. Diversifying the population helped to decrease the convergence rate with a chance of increasing accuracy.  The advantage of our proposed method becomes notable when the population is small. We can defer that our approach is also applicable to other real-time applications.
In summary, compared with the baseline, manual tuning methods, model-free and model-based HPO methods \cite{zahedi2020leveraging, zahedi2021search,zahedihyp} OptABC algorithm outperforms other methods in predicting student's success. OptABC uses an ABC-based algorithm to optimized the hyper-parameters of ML models, improves the classification accuracy effectually.

\section{Conclusion}
\label{sec:Conclusion}
We presented OptABC, an artificial bee colony (ABC)-based optimization
algorithm for hyper-parameter tuning for machine learning methods dealing with large datasets. Our proposed OptABC integrates ABC algorithm, K-Means clustering, greedy algorithm, and opposition-based learning strategy for tuning the hyper-parameters of different machine learning models.
We demonstrated OptABC in three experiments on SVM, RF and XGBoost. 

This hybrid method introduces a more intelligent approach to ABC and aimed to improve ABC's exploitation, exploration, and acceleration. The related experimental results on a real-world dataset compared with previous approaches on the same dataset demonstrate that the proposed OptABC exhibits faster convergence speed and running time without decreasing the accuracy in most cases and has an advantage over previous methods. In our future work, we plan to design strategies to further eliminate poor food sources in the initialization step to spend more of the evaluations on better solutions.
\bibliographystyle{IEEEtran}
\bibliography{ref}

\end{document}